\documentclass{article} 
\usepackage{iclr2023_conference,times}

\usepackage[colorlinks = true,
            linkcolor = red,
            urlcolor  = teal,
            citecolor = teal,
            anchorcolor = black,
            linktoc = all]{hyperref}
\usepackage{url}

\usepackage{wrapfig}
\usepackage{subfigure}
\usepackage{graphics,graphicx}
\usepackage{capt-of}
\usepackage{booktabs}
\usepackage{multirow}
\usepackage{array}
\usepackage{tabularx}

\usepackage{amsmath,amsfonts,bm}

\def\eqref#1{equation~\ref{#1}}

\def\1{\bm{1}}

\DeclareMathAlphabet{\mathsfit}{\encodingdefault}{\sfdefault}{m}{sl}
\SetMathAlphabet{\mathsfit}{bold}{\encodingdefault}{\sfdefault}{bx}{n}

\newcommand*{\tran}{^{\mkern-1.5mu\mathsf{T}}}
\newcolumntype{x}[1]{>{\centering\arraybackslash}p{#1}}
\renewcommand{\eqref}[1]{(\ref{#1})}

\title{Dataset Condensation with Latent Space \\ Knowledge Factorization and Sharing}

\author{%
  Hae Beom Lee$^{1*}$, Dong Bok Lee$^{1}$\thanks{: Equal Contribution}$\ \ $,$\ $ Sung Ju Hwang$^{1,2}$\\
    KAIST$^{1}$, AITRICS$^{2}$, South Korea \\
  \texttt{\{haebeom.lee, markhi, sjhwang82\}@kaist.ac.kr}  \\
}

\iclrfinalcopy 
\begin{document}

\maketitle

\begin{abstract}
In this paper, we introduce a novel approach for systematically solving dataset condensation problem in an efficient manner by exploiting the regularity in a given dataset. Instead of condensing the dataset directly in the original input space, we assume a generative process of the dataset with a set of learnable \emph{codes} defined in a compact latent space followed by a set of tiny \emph{decoders} which maps them differently to the original input space. By combining different codes and decoders interchangeably, we can dramatically increase the number of synthetic examples with essentially the same parameter count, because the latent space is much lower dimensional and since we can assume as many decoders as necessary to capture different styles represented in the dataset with negligible cost. Such knowledge factorization allows efficient sharing of information between synthetic examples in a systematic way, providing far better trade-off between compression ratio and quality of the generated examples. We experimentally show that our method achieves new state-of-the-art records by significant margins on various benchmark datasets
such as SVHN, CIFAR10, CIFAR100, and TinyImageNet.
\end{abstract}
\section{Introduction}

Deep learning has been successful in numerous machine learning problems thanks to the recent progress in parallel processing and the huge amount of real-world data collected from various sources. However, in some machine learning applications it is required to rehearse the training process repeatedly, such as hyperparameter optimization~\citep{bengio2000gradient,franceschi2017forward}, neural architecture search~\citep{liu2018darts} and continual learning~\citep{lopez2017gradient}. In those cases, it is prohibitive to keep and rehearse all the examples in huge datasets, giving rise to the need for compressing each dataset into a small set of representative examples. The conventional approaches resort to selecting a coreset~\citep{phillips2016coresets,toneva2018empirical,borsos2020coresets}, but their assumption is limited as there may not exist strongly representative examples in the original dataset~\citep{zhao2021datasetb} and also suffer from the difficulties of combinatorial optimizations~\citep{borsos2020coresets}. Recently, more popular approaches focus on \emph{dataset condensation}~\citep{wang2018dataset,zhao2021dataseta} which directly parameterize and optimize the synthetic dataset with gradient descent. They have shown to perform well compared to coreset selection approaches and generate plausibly looking examples as well. 

However, despite its great potential to other adjacent fields, most of the existing dataset condensation methodologies do not focus on exploiting the regularity in the dataset, such as what the underlying data generating process would be. They usually parameterize a set of synthetic examples directly in input space and minimize a distance between real and synthetic dataset in either parameter~\citep{zhao2021dataseta} or feature space~\citep{zhao2022dataset}. Such lack of assumptions on the underlying data generating process prevents them from efficiently sharing knowledge among the synthetic examples. Therefore, those approaches may require much larger space consumption to capture the same amount of information compared to when we can properly organize the data generating process. This intuition gives rise to questions about whether we can further improve the efficiency in solving the problem by exploiting the regularity in the given dataset.

\begin{figure}[t]
    \vspace{-0.2in}
	\centering
	\hfill
	\subfigure{
	    \includegraphics[height=2.5cm]{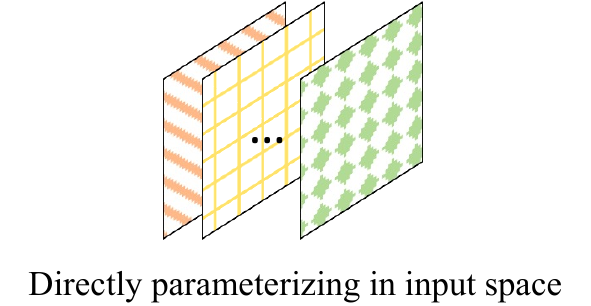}
	}
	\hfill
	\subfigure{
	    \includegraphics[height=2.5cm]{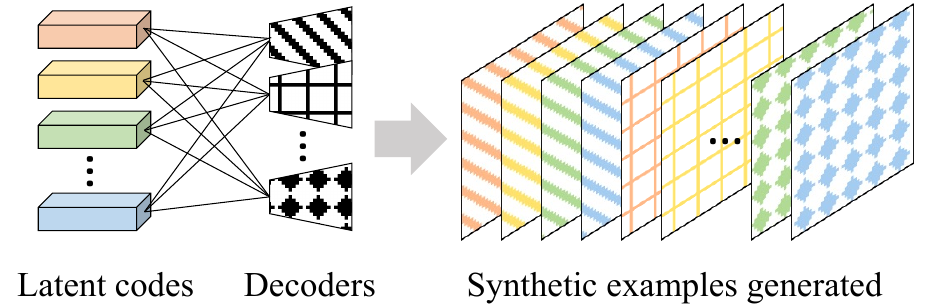}
	}
	\hfill
	\vspace{-0.1in}
	\caption{\small \textbf{(Left)} Conventional dataset condensation approaches that directly parameterize the synthetic examples in input space. \textbf{(Right)} Our approach that parameterize latent codes and multiple decoders that can generate much more number of examples with essentially the same parameter count. }
    \label{fig:concept}
    \vspace{-0.05in}
\end{figure}

To this end, we naturally assume that each datapoint is generated from a \emph{code} in a compact latent space followed by a sharable mapping function from that space to the original input space (i.e. \emph{decoder}), both of which are learned in an end-to-end manner. In this way, we can dramatically increase the number of generated synthetic examples with the same parameter count as the latent space can be much lower dimensional than the original input space, and also since we can assume as many decoders as appropriate to capturing various \emph{styles} represented in the dataset. See Figure~\ref{fig:concept} for the concept. Note that the additional per-class parameter count introduced by the decoders is negligible because those decoders are very tiny and can be shared across all the synthetic examples and classes. Furthermore, the knowledge is shared among the generated synthetic examples through the shared latent space and the set of shared decoders applicable to any latent codes interchangeably, dramatically improving the trade-off between compression ratio and quality of the synthetic dataset. To our knowledge, this paper is first to successfully employ such latent space approach to systematically solving the dataset condensation problem.

In algorithmic side, we show that the previous approaches that subsample synthetic examples for computational efficiency (e.g. gradient matching~\citep{zhao2021dataseta} or distribution matching~\citep{zhao2022dataset}) actually produce biased gradients, preventing the synthetic examples from being diversified. We further experimentally observe that subsampling of the real examples produce high-variance gradients, degrading the quality of the generated synthetic examples. Those observations indicate that we have to perform full batch training for both real and synthetic dataset. We thus adopt distribution matching~\citep{zhao2022dataset} as it does not require expensive second order computations unlike gradient matching~\citep{zhao2021dataseta}. We also prestore and reuse some of the information needed to perform distribution matching for further improving the training efficiency.

We experimentally show that our method, which we name as Knowledge Factorization and Sharing (KFS), achieves new state-of-the-art records on various benchmark datasets such as SVHN, CIFAR10, and CIFAR100 and TinyImageNet with different amount of parameter count allowed (e.g. $1$, $10$, or $50$ images per class). We also show that the synthetic examples generated from KFS produce decent performances on other network architectures than what they have been trained on (e.g. from ConvNet-3 to DenseNet-121), 
and it provides superior performance over various amount of training budget at evaluation time without overfitting.
We summarize the contribution of this paper as follows:
\vspace{-0.05in}
\begin{itemize}
    \item We introduce KFS, a novel approach for solving dataset condensation problem that factorizes the synthetic data generating process into a set of learnable latent codes and decoders, dramatically increasing the number of synthetic examples with a limited parameter count. 
    \item We show that the assumed generative process effectively allows knowledge sharing between the synthetic examples such that they achieve far better trade-off between the compression ratio versus the quality of the generated examples.
    \item We further show that the existing dataset condensation approaches suffer from the issue of biased and high-variance gradient, and experimentally verify that full batch training significantly improves the quality of the synthetic dataset.
    \item The proposed KFS achieves new state-of-the-art records in all datasets we considered, outperforming the previous state-of-the-art by significant margins.
\end{itemize}

\section{Related Work}

\vspace{-0.05in}
\paragraph{Dataset condensation.}
\cite{wang2018dataset} firstly introduced a differentiable approach to dataset condensation problem through a bilevel optimization formulation. However, in their work only a few inner-gradient steps are assumed for computational efficiency and additionally a shared initialization is learned and entangled with the learned data. \cite{zhao2021dataseta} overcome the limitation by trying to match the network weights and approximating the expensive bi-level optimization with short-horizon approximation, resulting in gradient matching between real and synthetic examples for each step. \cite{zhao2021datasetb} further develop the idea by learning a differentiable siamese networks for augmenting the synthetic examples. \cite{cazenavette2022dataset} recently propose to match much longer segments of learning trajectories, yielding competitive performances by guiding networks to a more similar state. On the other hand, instead of matching the network weights, \cite{zhao2022dataset} match distributions of real and synthetic dataset in a feature space. Although they sometimes show infereior performance to gradient matching, their approach is computationally more efficient as they do not require to compute expensive second-order derivatives such as vector-Jacobian products (VJPs) required by gradient matching. Simiilarly, \cite{wang2022cafe} propose to layer-wisely align features with auxiliary discriminative loss and bi-level formulation for adjusting parameter updates. On the other hand, \cite{nguyen2020dataset,nguyen2021dataset} propose to make use of the connection between infinitely wide neural networks and kernel ridge regression, but their approach requires hundreds of GPUs for training. Perhaps the most relevant to our work is recently proposed IDC~\citep{kim2022dataset}. They generate various synthetic data from a single set of condensed examples by segmenting and upscaling the condensed examples. Their approach can be seen as exploiting the regularity in the dataset, for instance, spatially nearby pixels look similar to each other. Although they achieve strong performances on many datasets, it is questionable whether the proposed generative process is sophisticated enough to fully exploit the regularity in the dataset. In this work, we make a more natural and general assumption on the data generating process, yielding new state-of-the-art records on all the datasets we consider.

\vspace{-0.05in}
\paragraph{Generative models.} Our work is closely related to generative models such as VAE~\citep{kingma2013auto} and GAN~\citep{goodfellow2014generative}. They maximize the likelihood of data by learning a compact latent space and a shared decoder, similarly to our work. However, it is questionable whether maximizing the data likelihood would be sufficient for the target task such as classification. Also, whereas it has been shown that modality is crucial for the performance of condensed dataset~\citep{kim2022dataset}, they are known to suffer from the posterior collapse (or mode collapse) problem~\citep{razavi2018preventing}. \cite{van2017neural} and \cite{razavi2019generating} overcome the issue by proposing VQ-VAE that models discrete representations in the latent space. The discreteness of the approach makes them closely related to the dataset condensation, 
but it is questionable whether such generative models can outperform the dataset condensation approaches recently introduced. For instance,
\cite{such20generative} combine GAN with meta-learning to generate synthetic examples, but they only slightly outperform~\cite{wang2018dataset}.

\vspace{-0.05in}
\paragraph{Object files and schemata.} Our approach is also closely related to object files and schemata discussed in \cite{goyal2020inductive,goyal2021factorizing,goyal2021neural}. Object files typically refer to the state of each object (e.g. states in RL or an object in an image) and schemata determine how they behave and update the object files correspondingly (e.g. actions in RL or styles in images). By carefully factorizing those two different types of knowledge, the knowledge structure becomes more efficient as we can apply the same schema to various objects, and more effectively generalizable to out-of-distribution because the architecture has better \emph{systematicity}. Also, the interactions between object files and schemata are inherently sparse, i.e. in real-world scenarios, explaining a phenomenon usually requires only a few rules given a set of objects. In our case, a single pair of object file and schema is assumed to be sufficient to generate a single synthetic image. 
\vspace{-0.05in}
\section{Approach}
\vspace{-0.05in}
We next present our approach, Knowledge Factorization and Sharing (KFS), that efficiently factorize the data generating process into latent codes and decoders so that knowledge is effectively shared among the synthetic examples generated.
\vspace{-0.05in}
\subsection{Knowledge Factorization and Sharing in Latent Space}
\label{section:kfs}
\vspace{-0.05in}
It was observed by \citet{kim2022dataset} that what determines the performance of condensed dataset is its modality rather than resolution. We follow the same principle but we want to find even better trade-off between increasing the modality of the dataset and preserving or even improving the quality of generated images. 

\paragraph{Latent code - decoder factorization.}
Our approach starts from assuming that the datapoints lie in a compact latent space whose dimension can be much lower than that of the actual input space. Thanks to its lower dimensionality, we can dramatically increase the number of datapoints given a limited parameter count (e.g. the number of parameters used by synthetic images per each class). We can also preserve the quality of the synthetic images to the extent to which the assumption holds, which has been proven to be generally the case. The only space overhead is additional parameters for a tiny decoder that maps those latent codes into the input space. Note that this decoder can be shared across all the datapoints and classes, and even if the decoder is relatively big we can flexibly adjust the number of latent codes so that the total parameter counts are comparable. 

Let us take an illustrative example. Suppose we want to condense a dataset consisting of images of shape $3 \times 32 \times 32 = 3072$ (e.g. SVHN or CIFAR10). Instead of learning those $3072$ parameters in the pixel space, we instead learn much smaller codes in, say $12 \times 4 \times 4 = 192$ dimensional latent space and decode them back to the original pixel space with a shared decoder. Given the same parameter count, this allows us to generate $16$ times more number of images than the existing approaches ($192 \times 16 = 3072$). Note that the per-class parameter count of shared decoder is negligible. In the above example, we could use a tiny 3-layer deconvolutional architecture with channels decreasing as $12 \rightarrow 9 \rightarrow 6 \rightarrow 3$ and $2 \times 2$ kernel, resulting in only $738$ parameters in total. Since this decoder can be shared across all classes, dividing it with the number of classes (e.g. $10$) results in a negligible amount of per-class parameter count (e.g. $73.8$) compared to the actual image size.

\paragraph{Sharing various styles with multiple decoders.}

The above factorization means that now we decouple the compressed expression of the instances from how to decode them back. This decoding process can be thought of as adding details to the latent codes so that the decoded instances have sufficient information in the original input space. For instance, given a compressed expression of a car, the decoder may add in or supplement background details or colors to complete an image. 

This insight implies that the number of decoders can be as many as how various \emph{styles} exists in the actual dataset. For instance, given the same compressed expression of an airplane, the background may be ocean, sky, forest, or sunset. By assuming various decoders, we could expect each of the decoders to explain each of those distinct styles in the dataset respectively. Furthermore, since the knowledge about how to add styles to the latent codes can be shared across all instances and classes, we can dramatically increase the number of synthetic images without increasing parameter count too much, similarly to cartesian product between a pair of sets. For instance, if there are $8$ shared decoders and $16$ latent codes for each of $10$ classes, we can generate total $8 \times 16 = 128$ synthetic images with only around $3131$ parameter count for each class, which is less than $2\%$ increase from the parameter count of a single image $3 \times 32 \times 32 = 3072$.

\subsection{Training with Unbiased and Low-variance Distribution Matching}

Basically, our approach is compatible with any learning-based dataset condensation approaches that allows to differentiate w.r.t latent codes and decoder parameters by backpropagating through the generated examples, such as gradient matching~\citep{zhao2021dataseta} or distribution matching~\citep{zhao2022dataset}. In this paper, we propose to use distribution matching as it does not require expensive vector-Jacobian products (VJPs) computations whereas gradient matching requires to compute VJP for every training step.

\paragraph{Distribution matching.} Here we introduce notations for further discussion. We denote a set of $N$ real examples for each class $c=1,\dots,C$ as $\mathcal{X}_c = \{x_{c,1},\dots,x_{c,N}\}$. Similarly, a set of $M$ latent codes for each class $c$ is $\Theta_c = \{\theta_{c,1},\dots,\theta_{c,M}\}$ and we collect them into $\Theta = \{\Theta_1,\dots,\Theta_C\}$. Note that the dimension of $\theta$ is much lower than that of $x$. $g(x)$ is a shared feature extractor taking $x$ as an input. $f(\theta;\phi_d)$ is $d$-th decoder parameterized by $\phi_d$ that takes a latent code $\theta$ as an input. We assume there are $D$ decoders and collect their parameters into $\phi = \{\phi_1,\dots,\phi_D\}$. With these notations, we consider the following empirical MMD loss~\citep{gretton2012kernel,zhao2022dataset}.
\begin{align}
     L(\Theta, \phi) = \frac{1}{C}\sum_{c=1}^C \frac{1}{2}\left\| \frac{1}{N} \sum_{n=1}^{N} g(x_{c,n}) - \frac{1}{DM}\sum_{d=1}^D \sum_{m=1}^{M} g(f(\theta_{c,m};\phi_d)) \right\|_2^2
     \label{eq:dm}
\end{align}
Following \cite{zhao2022dataset}, we sample $g(\cdot)$ from random initializations rather than from a set of pretrained networks, which is computationally more efficient and works well in practice. 

\paragraph{Bias due to subsampling of synthetic examples.} \cite{zhao2022dataset} further approximate the first term (i.e. mean of $g(x)$ over $n$) and the second term (i.e. mean of $g(f(\theta;\phi))$ over $d$ and $m$) by random subsampling for computational efficiency. However, whereas the gradient $\nabla_{\Theta,\phi}L(\Theta,\phi)$ is unbiased w.r.t the random subsampling of indices of real data $\{n\}_{n=1}^N$ or classes $\{c\}_{c=1}^C$, the gradient is actually biased w.r.t the random subsampling of indices of latent codes $\{m\}_{m=1}^M$ or decoders $\{d\}_{d=1}^D$. Specifically, suppose for simplicity we randomly sample a single pair of $m \sim \text{Uniform}(1,M)$ and $d \sim \text{Uniform}(1,D)$ and denote the corresponding loss as $\hat{L}$. Then we can easily compute the bias of gradient $\mathbb{E}_{m,d}[\nabla_{\Theta,\phi}\hat{L}(\Theta,\phi)] - \nabla_{\Theta,\phi} L(\Theta,\phi)$ as
\begin{equation}
\begin{aligned}
   \nabla_{\Theta, \phi} \frac{1}{C}\sum_{c=1}^C \frac{1}{2} \Bigg\{
    &\frac{1}{DM} \sum_{d=1}^D \sum_{m=1}^M g(f(\theta_{c,m};\phi_d))\tran g(f(\theta_{c,m};\phi_d)) 
    \\
    &- \frac{1}{D^2M^2} \sum_{d=1}^D\sum_{d'=1}^D\sum_{m=1}^M\sum_{m'=1}^M g(f(\theta_{c,m};\phi_{d}))\tran g(f(\theta_{c,m'};\phi_{d'}))
    \Bigg\}.
    \label{eq:bias}
\end{aligned}
\end{equation}
See Appendix~\ref{sec:bias} for the derivation. The above result suggests that the bias comes from ignoring the interactions within the latent codes and within the decoders, respectively. The biased gradient fails to minimize the inner product of the embeddings computed from different codes and decoders, such as $g(f(\theta_{c,m};\phi))\tran g(f(\theta_{c,m'};\phi))$ and $g(f(\theta_{c,m};\phi_d))\tran g(f(\theta_{c,m};\phi_{d'}))$, preventing the codes and decoders from being diversified. Intuitively, if we sample $\theta$, then each individual $\theta \in \Theta_c$ will try to explain the whole class set $\mathcal{X}_c$ at a time without considering the existence of other codes $\theta' \in \Theta_c \backslash \{\theta\}$ due to the sampling. The same intuition holds for sampling the decoder $f(\cdot;\phi_d)$. Therefore, we should not sample $m$ and $d$. However, despite of its importance, most of the recent works do not consider correcting the bias~\citep{zhao2021dataseta,zhao2021datasetb,zhao2022dataset,  cazenavette2022dataset,wang2022cafe,kim2022dataset}.

\vspace{-0.05in}
\paragraph{Variance due to subsampling of real examples.} We also found that random subsampling of real examples is detrimental to performance due to large variance. Again, suppose for simplicity we randomly sample a single index $n \sim \text{Uniform}(1,N)$ independently for each class $c$ and denote the corresponding loss as $\tilde{L}$. Then the variance of the gradient $\text{Var}_n(\nabla_{\Theta,\phi}\tilde{L}(\Theta,\phi))$ is 
\begin{align}
    \frac{1}{C^2}\sum_{c=1}^C V_c\tran \left\{
    \frac{1}{N} \sum_{n=1}^N g(x_{c,n}) g(x_{c,n} )\tran
    - \left(\frac{1}{N} \sum_{n=1}^N  g(x_{c,n})\right) \left(\frac{1}{N} \sum_{n=1}^N  g(x_{c,n})\right)\tran
    \right\} V_c
    \label{eq:variance}
\end{align}
where $V_c = \frac{1}{DM}\sum_{d=1}^d\sum_{m=1}^M\nabla_{\Theta,\phi}g(f(\theta_{c,m};\phi_d))$. See Appendix~\ref{sec:variance} for the derivation. The middle term in the big braces is nothing but the variance of embeddings of real examples $g(x_{c,n})$ over $n=1,\dots,N$. The problem is that since we sample $g(\cdot)$ from random initializations according to the original distribution matching algorithm~\citep{zhao2022dataset}, the variance of $g(x_{c,n})$ tends to be very high compared to the discriminative power of the representation $g(\cdot)$ across the classes. We empirically observe that such relatively high variance makes the gradient $\nabla_{\Theta,\phi}\tilde{L}(\Theta,\phi)$ too noisy for the training process to find a good solution.

\vspace{-0.05in}
\paragraph{Full batch training.} Based on the above analysis, we propose to perform full batch training for both real and synthetic dataset for unbiased and low-variance training. In case of real dataset, we basically need to compute the embedding mean of real examples $\frac{1}{N}\sum_{n=1}^N g(x_{c,n})$ across the classes and random weights $w_1,\dots,w_K$, with $K$ denoting the total training steps.  Instead of repeatedly computing them for every experiment, we speed up our experiments by precomputing those means once and storing them in an external device with the corresponding random weights, so that we can quickly restore them for the later experiments. In case of synthetic dataset, we simply distribute the classes over multiple GPUs and accumulate the gradients over all the processes. When the size of each class is too large to fit in a single GPU, we compute the embedding mean over the whole synthetic examples but backpropagate through only a subset of them for handling memory issue, while keeping the gradient unbiased.

\vspace{-0.05in}
\paragraph{Discussion.} 
Although the distribution matching objective in Eq.~\eqref{eq:dm} works well in practice, it is not clear why it should work with repeated sampling of $g(\cdot)$ from random initializations. A simple explanation may be that for each class, a well-condensed synthetic dataset should lead to a similar embedding mean to that of real dataset regardless of those random initializations. However, further justification and understanding of the objective is required, which we leave as a future work.

\begin{table}[t!]
    \vspace{-0.1in}
	\caption{\small \textbf{Classification accuracies (\%) on ConvNet-3.} For our method, we report mean and standard deviation over 15 evaluations (3 training runs and 5 evaluations for each run).} \label{tbl:low_resolution}
	\vspace{-0.15in}
	\begin{center}
	    \small
        \resizebox{1.\linewidth}{!}{
		\begin{tabular}{c|ccc|ccc|cc|cc}
			\toprule
			  Dataset &  \multicolumn{3}{c|}{SVHN} &  \multicolumn{3}{c|}{CIFAR10} & \multicolumn{2}{c|}{CIFAR100} & \multicolumn{2}{c}{TinyImageNet}  \\
			  Images / Class & 1 & 10 & 50 & 1 & 10 & 50 & 1 & 10 & 1 & 10 \\
			  Param. / Class & 3072 & 30720 & 153600 & 3072 & 30720 & 153600 & 3072 & 30720 & 12288 & 122880  \\
			\hline
			\hline
			Random
			& 14.6\tiny$\pm$1.6
			& 35.1\tiny$\pm$4.1
			& 70.9\tiny$\pm$0.9
			& 14.4\tiny$\pm$2.0
			& 26.0\tiny$\pm$1.2
			& 43.4\tiny$\pm$1.0
			& 4.2\tiny$\pm$0.3
			& 14.6\tiny$\pm$0.5
			& 1.4\tiny$\pm$0.1
			& 5.0\tiny$\pm$0.2
			\\
			Herding
			& 20.9\tiny$\pm$1.3
			& 50.5\tiny$\pm$3.3
			& 72.6\tiny$\pm$0.8
			& 21.5\tiny$\pm$1.2
			& 31.6\tiny$\pm$0.7
			& 40.4\tiny$\pm$0.6
			& 8.4\tiny$\pm$0.3
			& 17.3\tiny$\pm$0.3
			& 2.8\tiny$\pm$0.2
			& 6.3\tiny$\pm$0.2
			\\
			\hline
			DC
			& 31.2\tiny$\pm$1.4
			& 76.1\tiny$\pm$0.6
			& 82.3\tiny$\pm$0.3
			& 28.3\tiny$\pm$0.5
			& 44.9\tiny$\pm$0.5
			& 53.9\tiny$\pm$0.5
			& 12.8\tiny$\pm$0.3
			& 25.2\tiny$\pm$0.3
			& -
			& -
			\\
			DSA
			& 27.5\tiny$\pm$1.4
			& 79.2\tiny$\pm$0.5
			& 84.4\tiny$\pm$0.4
			& 28.8\tiny$\pm$0.7
			& 52.1\tiny$\pm$0.5
			& 60.6\tiny$\pm$0.5
			& 13.9\tiny$\pm$0.3
			& 32.3\tiny$\pm$0.3
			& -
			& -
			\\
			DM
			& 20.3\tiny$\pm$2.1
			& 73.5\tiny$\pm$1.0
			& 84.2\tiny$\pm$0.0
			& 26.0\tiny$\pm$0.8
			& 48.9\tiny$\pm$0.6
			& 63.0\tiny$\pm$0.4
			& 11.4\tiny$\pm$0.3
			& 29.7\tiny$\pm$0.3
			& 3.9\tiny$\pm$0.2
			& 12.9\tiny$\pm$0.4
			\\
			KIP to NN
			& 57.3\tiny$\pm$0.1
			& 75.0\tiny$\pm$0.1
			& 80.5\tiny$\pm$0.1
			& 49.9\tiny$\pm$0.2
			& 62.7\tiny$\pm$0.3
			& 68.6\tiny$\pm$0.2
			& 15.7\tiny$\pm$0.2 
			& 28.3\tiny$\pm$0.1
			& -
			& -
			\\
			CAFE + DSA
			& 42.9\tiny$\pm$3.0
			& 77.9\tiny$\pm$0.6
			& 82.3\tiny$\pm$0.4
			& 31.6\tiny$\pm$0.8
			& 50.9\tiny$\pm$0.5
			& 62.3\tiny$\pm$0.4
			& 14.0\tiny$\pm$0.3
			& 31.5\tiny$\pm$0.2
			& -
			& -
			\\
			Traj. Matching
			& -
			& -
			& -
			& 46.3\tiny$\pm$0.8
			& 65.3\tiny$\pm$0.7
			& 71.6\tiny$\pm$0.2
			& 24.3\tiny$\pm$0.3
			& 40.1\tiny$\pm$0.4
			& 8.8\tiny$\pm$0.3
			& 23.2\tiny$\pm$0.2
			\\
			IDC
			& 68.1\tiny$\pm$0.1
			& 87.3\tiny$\pm$0.2
			& 90.2\tiny$\pm$0.1
			& 50.0\tiny$\pm$0.4
			& 67.5\tiny$\pm$0.5
			& 74.5\tiny$\pm$0.1
			& -
			& 44.8\tiny$\pm$0.2
			& -
			& -
			\\
			\textbf{KFS (ours)}
			& \textbf{82.9\tiny$\pm$0.4}
			& \textbf{91.4\tiny$\pm$0.2}
			& \textbf{92.2\tiny$\pm$0.1}
			& \textbf{59.8\tiny$\pm$0.5}
			& \textbf{72.0\tiny$\pm$0.3}
			& \textbf{75.0\tiny$\pm$0.2}
			& \textbf{40.0\tiny$\pm$0.5}
			& \textbf{50.6\tiny$\pm$0.2}
			& \textbf{22.7\tiny$\pm$0.2}
			& \textbf{27.8\tiny$\pm$0.2}
			\\
			\hline
			Full dataset
			& \multicolumn{3}{c|}{95.4\tiny$\pm$0.1} 
			& \multicolumn{3}{c|}{84.8\tiny$\pm$0.1} 
			& \multicolumn{2}{c|}{56.2\tiny$\pm$0.3}
    		& \multicolumn{2}{c}{37.6\tiny$\pm$0.4} 
			\\
			\bottomrule
		\end{tabular}
		}
	\end{center}
	\small
	\vspace{-0.25in}
\end{table}
\begin{table}[t!]
	\caption{\small \textbf{Cross architecture experiments.} Conv3, RN10, and DN121 denote ConvNet-3, ResNet-10, and DenseNet-121, respectively. We train on ConvNet-3 and evaluate on the three architectures.} \label{tbl:cross_architecture}
	\vspace{-0.15in}
	\begin{center}
	    \small
        \resizebox{\linewidth}{!}{
		\begin{tabular}{c|c|ccc|ccc|ccc}
			\toprule
			 \multirow{2}{*}{Dataset} & Images / Class &  \multicolumn{3}{c|}{1} &  \multicolumn{3}{c|}{10} & \multicolumn{3}{c}{50} \\
			  & Test Architecture & Conv3 & RN10 & DN121 & Conv3 & RN10 & DN121 & Conv3 & RN10 & DN121 \\
			\hline
			\hline
			\multirow{5}{*}{SVHN}
			& DSA
			& 27.5\tiny$\pm$1.4
			& 13.2\tiny$\pm$1.1
			& 13.3\tiny$\pm$1.4
			& 79.2\tiny$\pm$0.5
			& 19.5\tiny$\pm$1.5
			& 23.1\tiny$\pm$1.9
			& 84.4\tiny$\pm$0.4
			& 41.6\tiny$\pm$2.1
			& 58.0\tiny$\pm$3.1
			\\
			& DM
			& 20.3\tiny$\pm$2.1
			& 10.5\tiny$\pm$2.8
			& 13.6\tiny$\pm$1.0
			& 73.5\tiny$\pm$1.0
			& 28.2\tiny$\pm$1.5
			& 24.8\tiny$\pm$2.5
			& 84.2\tiny$\pm$0.0
			& 54.7\tiny$\pm$1.3
			& 58.4\tiny$\pm$2.7
			\\
			& IDC
			& 68.1\tiny$\pm$0.1
			& 39.6\tiny$\pm$1.5
			& 39.9\tiny$\pm$2.9
			& 87.3\tiny$\pm$0.2
			& 83.3\tiny$\pm$0.2
			& 82.8\tiny$\pm$0.2
			& 90.2\tiny$\pm$0.1
			& 89.1\tiny$\pm$0.2
			& \textbf{91.0\tiny$\pm$0.3}
			\\
			& \textbf{KFS (ours)}
			& \textbf{82.9\tiny$\pm$0.4}
			& \textbf{75.7\tiny$\pm$0.8}
			& \textbf{81.0\tiny$\pm$0.7}
			& \textbf{91.4\tiny$\pm$0.2}
			& \textbf{90.3\tiny$\pm$0.2}
			& \textbf{89.7\tiny$\pm$0.2}
			& \textbf{92.2\tiny$\pm$0.1}
			& \textbf{90.9\tiny$\pm$0.2}
			& 90.2\tiny$\pm$0.2
			\\
			\cline{2-11}
			& Full dataset
			& 95.4\tiny$\pm$0.1
			& 93.8\tiny$\pm$0.5
			& 89.1\tiny$\pm$0.8
			& 95.4\tiny$\pm$0.1
			& 93.8\tiny$\pm$0.5
			& 89.1\tiny$\pm$0.8
			& 95.4\tiny$\pm$0.1
			& 93.8\tiny$\pm$0.5
			& 89.1\tiny$\pm$0.8
			\\
			\hline
			\multirow{5}{*}{CIFAR10}
			& DSA
			& 28.8\tiny$\pm$0.7
			& 25.1\tiny$\pm$0.8
			& 25.9\tiny$\pm$1.8
			& 52.1\tiny$\pm$0.5
			& 31.4\tiny$\pm$0.9
			& 32.9\tiny$\pm$1.0
			& 60.6\tiny$\pm$0.5
			& 49.0\tiny$\pm$0.7
			& 53.4\tiny$\pm$0.8
			\\
			& DM
			& 26.0\tiny$\pm$0.8
			& 13.7\tiny$\pm$1.6
			& 12.9\tiny$\pm$1.8
			& 48.9\tiny$\pm$0.6
			& 31.7\tiny$\pm$1.1
			& 32.2\tiny$\pm$0.8
			& 63.0\tiny$\pm$0.4
			& 49.1\tiny$\pm$0.7
			& 53.7\tiny$\pm$0.7
			\\
			& IDC
			& 50.0\tiny$\pm$0.4
			& 41.9\tiny$\pm$0.6
			& 39.8\tiny$\pm$1.2
			& 67.5\tiny$\pm$0.5
			& 63.5\tiny$\pm$0.1
			& 61.6\tiny$\pm$0.6
			& 74.5\tiny$\pm$0.1
			& 72.4\tiny$\pm$0.5
			& 71.8\tiny$\pm$0.6
			\\
			& \textbf{KFS (ours)}
			& \textbf{59.8\tiny$\pm$0.5}
			& \textbf{47.0\tiny$\pm$0.8}
			& \textbf{49.5\tiny$\pm$1.3}
			& \textbf{72.0\tiny$\pm$0.3}
			& \textbf{70.3\tiny$\pm$0.3}
			& \textbf{71.4\tiny$\pm$0.4}
			& \textbf{75.0\tiny$\pm$0.2}
			& \textbf{75.1\tiny$\pm$0.3}
			& \textbf{76.3\tiny$\pm$0.4}
			\\
			\cline{2-11}
			& Full dataset
			& 84.8\tiny$\pm$0.1 
			& 87.9\tiny$\pm$0.2 
			& 90.5\tiny$\pm$0.3
			& 84.8\tiny$\pm$0.1 
			& 87.9\tiny$\pm$0.2 
			& 90.5\tiny$\pm$0.3 
			& 84.8\tiny$\pm$0.1  
			& 87.9\tiny$\pm$0.2  
			& 90.5\tiny$\pm$0.3
			\\
			\bottomrule
		\end{tabular}
		}
	\end{center}
	\vspace{-0.15in}
	\small
\end{table}
\begin{figure}[h]
	\centering
	\hskip -0.1in
	\subfigure{
	    \includegraphics[width=0.245\textwidth]{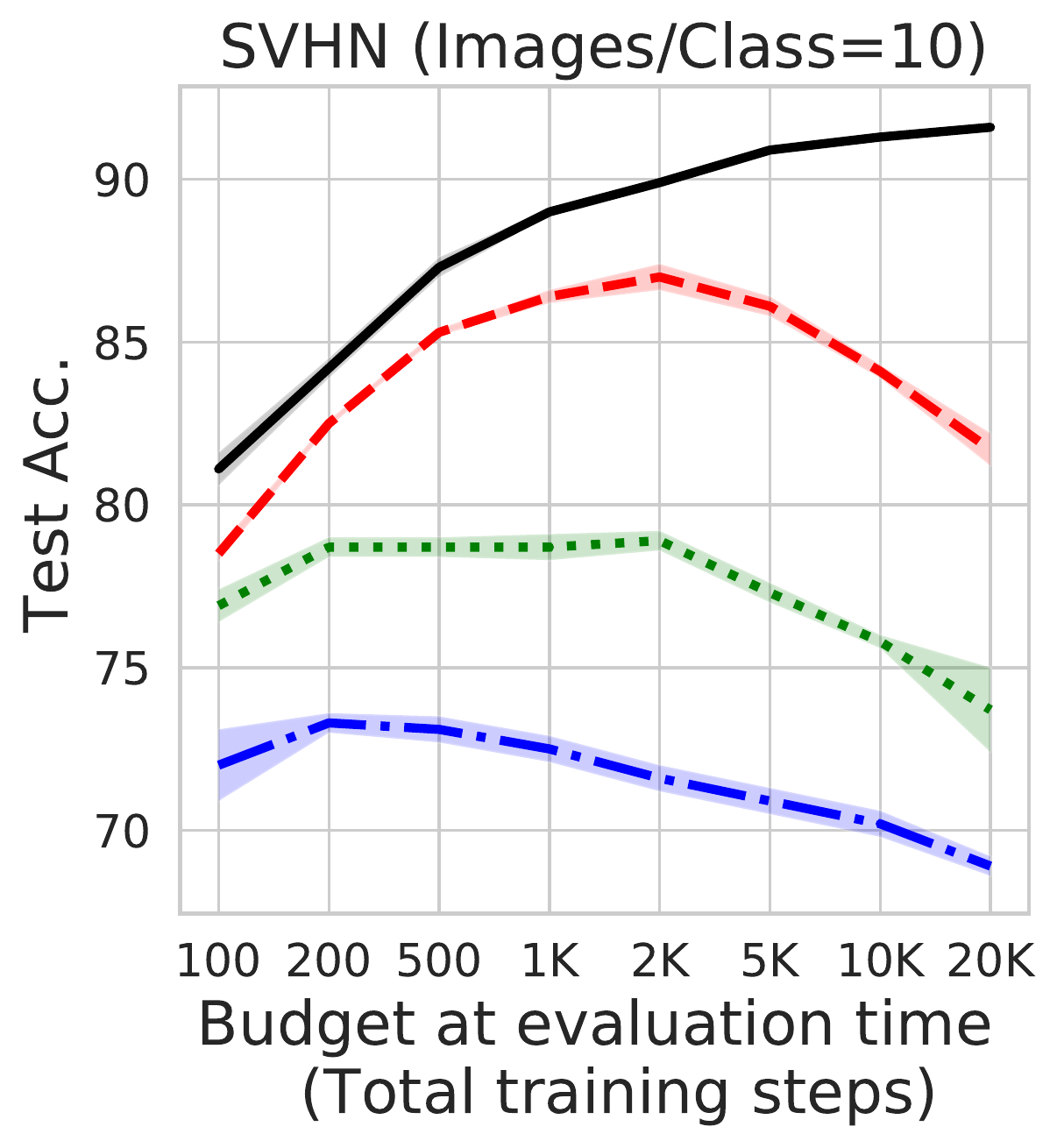}
	}
	\hskip -0.1in
	\subfigure{
	    \includegraphics[width=0.245\textwidth]{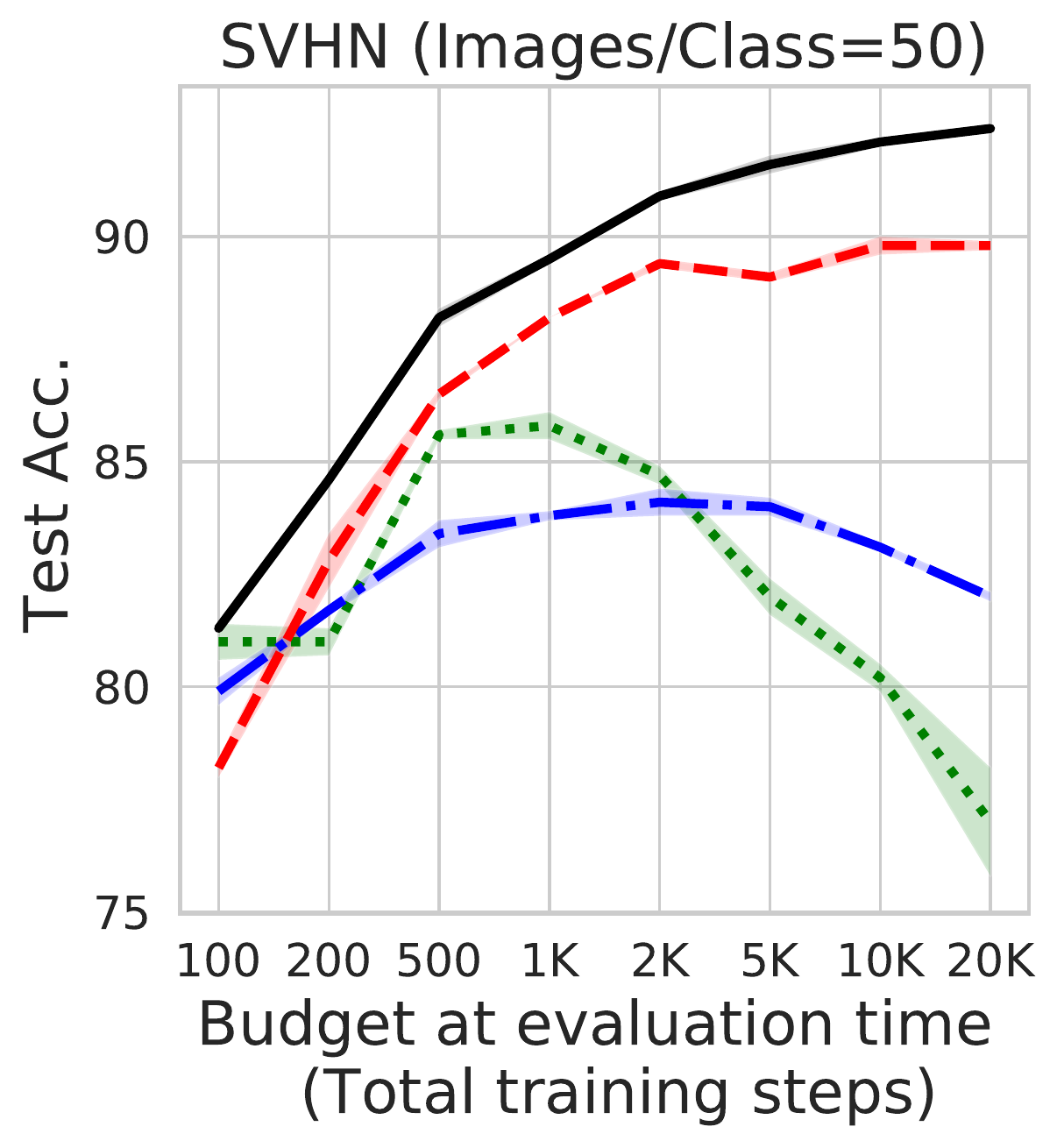}
	}
	\hskip -0.1in
	\subfigure{
	    \includegraphics[width=0.245\textwidth]{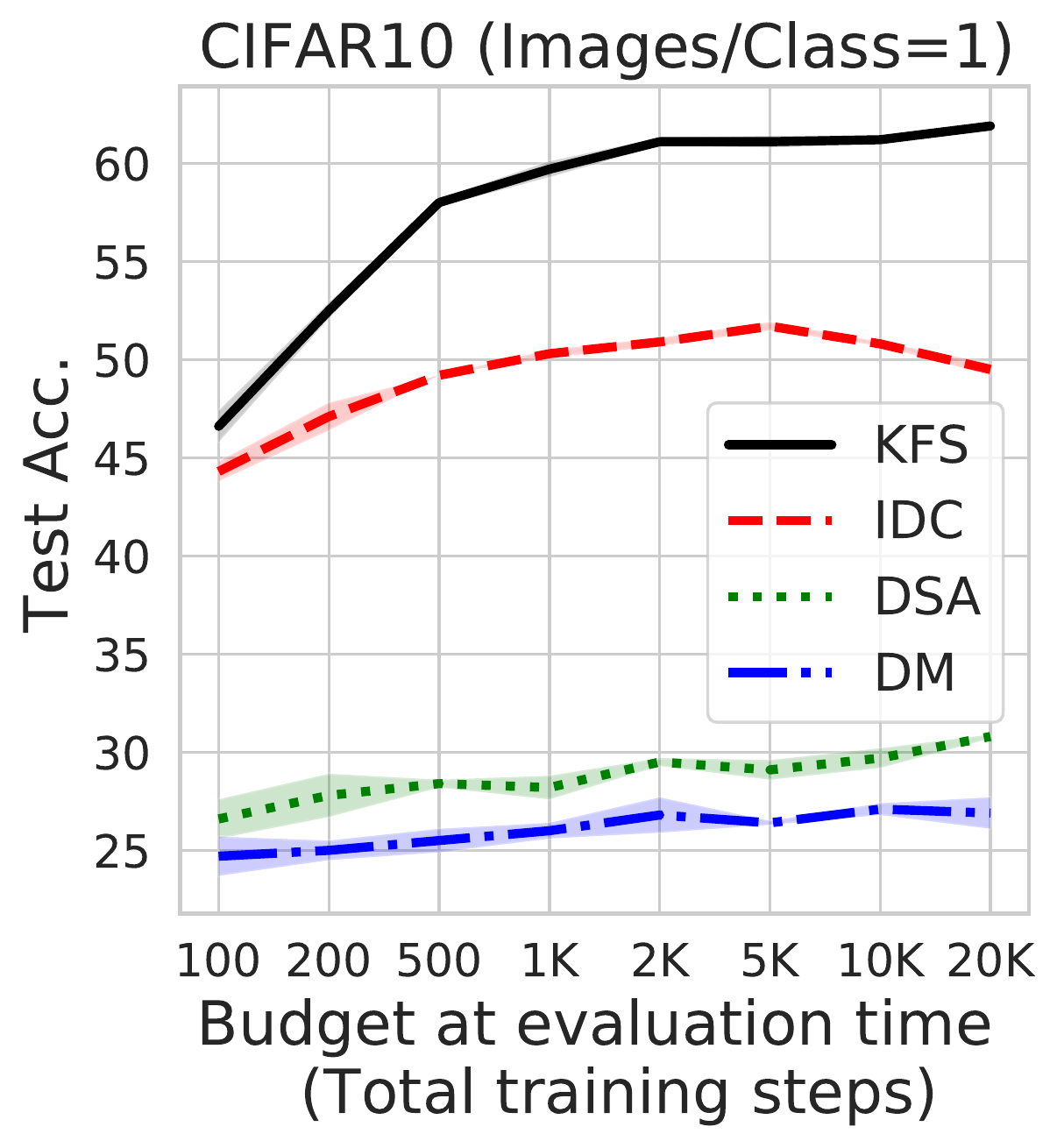}
	}
	\hskip -0.1in
	\subfigure{
	    \includegraphics[width=0.245\textwidth]{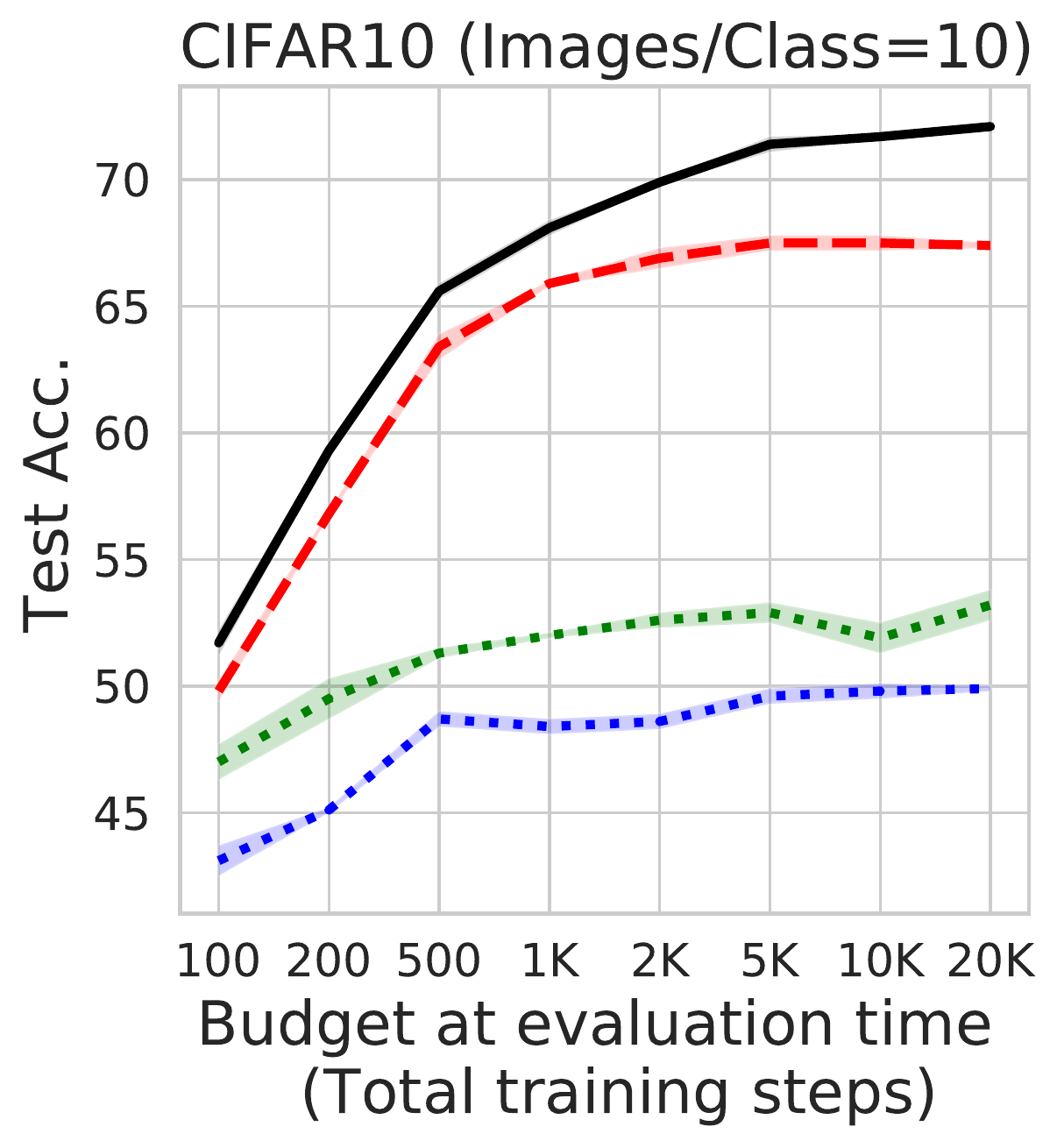}
	}
	\hskip -0.1in
	\vspace{-0.17in}
	\caption{\small The evaluation results across various amount of budgets defined as the number of training steps allowed for each evaluation. Batchsize is set to $\min(N_\text{total}, 256)$ where $N_\text{total}$ is the number of all examples in each synthetic dataset. }
    \label{fig:evaluation}
	\vspace{-0.15in}
\end{figure} 

\begin{figure}[t]
	\centering
	\hfill
	\subfigure{
	    \includegraphics[height=3.2cm]{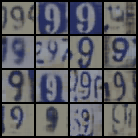}
	}
	\hfill
	\subfigure{
	    \includegraphics[height=3.2cm]{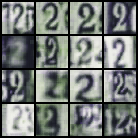}
	}
	\hfill
	\subfigure{
	    \includegraphics[height=3.2cm]{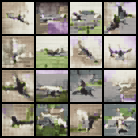}
	}
	\hfill
	\subfigure{
	    \includegraphics[height=3.2cm]{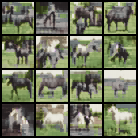}
	}
	\hfill	
	\subfigure{
	    \includegraphics[height=3.2cm]{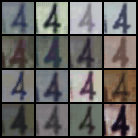}
	}
	\hfill
	\subfigure{
	    \includegraphics[height=3.2cm]{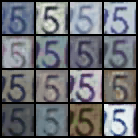}
	}
	\hfill
	\subfigure{
	    \includegraphics[height=3.2cm]{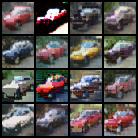}
	}
	\hfill
	\subfigure{
	    \includegraphics[height=3.2cm]{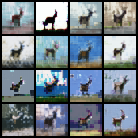}
	}
	\hfill
	\vspace{-0.15in}
	\caption{\small \textbf{(Top row)} Synthetic images (SVHN and CIFAR10) generated by varying latent codes while fixing a decoder. \textbf{(Bottom row)} Synthetic images generated by varying decoders while fixing a code. }
    \label{fig:mainfig}
	\vspace{-0.1in}
\end{figure} 

\vspace{-0.05in}
\section{Experiments}
\vspace{-0.05in}
We next experimentally validate the efficacy of KFS on various datasets and learning scenarios.


\textbf{Datasets. } We consider the following benchmark datasets such as SVHN  (32$\times$32)~\citep{netzer2011reading}, CIFAR10 (32$\times$32)~\citep{krizhevsky2009learning}, CIFAR100 (32$\times$32)~\citep{krizhevsky2009learning}, and TinyImageNet (64$\times$64)~\citep{le2015tiny}.
\textbf{Baselines. } We consider coreset selection approaches such as Random and Herding. For dataset condensation approaches, we compare with gradient (or trajectory) matching such as DC~\citep{zhao2021dataseta}, DSA~\citep{zhao2021datasetb}, Trajectory Matcing~\citep{cazenavette2022dataset}, and IDC~\citep{kim2022dataset}, as well as distribution matching such as DM~\citep{zhao2022dataset} and CAFE~\citep{wang2022cafe}. We also compare against KIP~\citep{nguyen2021dataset}. \textbf{Experimental setup. } For fair comparison, we use essentially the same parameter count to the baselines. For instance, if there are 10 classes with 10 ``Images / Class" (in Table~\ref{tbl:low_resolution} and Table~\ref{tbl:cross_architecture}) and the image shape is 3$\times$32$\times$32, then the total per-class parameter count is $30720$. In this case our KFS assumes 16 codes of shape 12$\times$4$\times$4 with the 8 decoders requiring only 73.8 additional parameters per class (see section~\ref{section:kfs}). See Appendix~\ref{sec:experimental_setup} for further details about experimental setup. 

\vspace{-0.05in}
\paragraph{Quantitative analysis.}
Table~\ref{tbl:low_resolution} shows the performance of the baselines and our KFS from each dataset with varying number of images (or parameter count) per class. We can see that KFS outperforms all the previous methods by significant margins in all datasets and settings. KFS especially performs well when the given parameter count per class is small. For instance, when ``Images / Class" $= 1$, KFS shows 14.8\% and 9.8\% performance improvements over IDC on SVHN and CIFAR10 respectively, the previous state-of-the-art recently introduced by~\cite{kim2022dataset}. This reconfirms the observation from~\cite{kim2022dataset} that increasing the modality in the synthetic dataset is more important than polishing a few high-quality examples when the parameter count is very limited. However, KFS provides much higher modality (i.e. more number of synthetic examples) than IDS with essentially the same parameter count because KFS can compress each instance far more tightly by systematically factorizing and sharing the knowledge across the synthetic examples.

\vspace{-0.05in}
\paragraph{Cross-architecture generalization.} Table~\ref{tbl:cross_architecture} shows how the baselines and KFS perform when the network architecture becomes different at the evaluation time (ResNet-10 and DenseNet-121) from the one that used for condensation (ConvNet-3). We can see that the performance of KFS is more robust to the change of network architectures. For instance, when ``Images/Class" $=$ 1 in SVHN dataset, the performance drop of KFS is only 7.2\% and 1.9\% for ResNet-10 and DenseNet-121 respectively, whereas the drop is 28.5\% and 28.2\% for IDC. Such robustness means that the learned synthetic examples from KFS are more natural as they are less specifically tailored to a specific network architecture. Similarly, when ``Images/Class" $=$ 50 in CIFAR10 dataset, the performance of KFS even improves as the architecture changes, while the performance of other baselines decreases.

\vspace{-0.05in}
\paragraph{Training budgets at evaluation time.}
One may wonder whether the good performance of KFS is solely due to the larger size of the synthetic dataset rather than the quality of each example, such that KFS inherently requires much longer evaluation time than the baselines. To address this question, we consider various amount of budget at evaluation time defined as total training steps allowed. In Figure~\ref{fig:evaluation} we can clearly see that KFS provides significantly better performances than the baselines across all range of budgets: from the limited budgets (100-500 training steps) to the longer training steps (1K-20K steps), demonstrating that the quality of synthetic examples generated from KFS is also comparable to or even better than that of the baselines. Interestingly, whereas the baselines often suffer from overfitting (e.g., SVHN dataset), our KFS has no such issue hence can provide more diverse options to users about how much longer to train their models for better performance.

\vspace{-0.05in}
\paragraph{Qualitative analysis.}
We next visualize the synthetic images generated from KFS in Figure~\ref{fig:mainfig}. We can see from the top row that KFS can generate various modes of the data distribution by varing the learned latent codes while fixing a decoder. For instance, in SVHN dataset the digits show various styles with different adjacent numbers, and in CIFAR10 dataset the orientations and shapes of objects are diverse, demonstrating the ability of the latent codes to capture multi-modality of each data distribution. We also fix a code and vary the decoders in the bottom row. The synthetic examples show various backgrounds and styles of the objects such as strokes and colors. Overall, the visualization clearly shows how the factorization between latent codes and decoders allows KFS to efficiently share knowledge between the synthetic examples.




\section{conclusion}
In this paper, we introduced a novel method for solving dataset condensation problem in a systematic and efficient way. Instead of parameterizing synthetic examples directly in input space, we proposed to fully exploit the regularity in a given dataset, such that we assume a generative process of synthetic examples based on the factorization between latent codes and decoders. Based on it, we showed how to increase the number of synthetic examples with essentially the same parameter count by combining the codes and decoders interchangeably. We further demonstrated how the knowledge is efficiently shared between synthetic examples with the framework, achieving superior trade-off between compression ratio and quality of the synthetic examples. Possible future directions include scaling up the method to achieve or even surpass the performances of the full dataset, and further developing the idea to multi-task or meta-learning setting where the goal is to share knowledge about how to compress a dataset between multiple tasks.

\bibliography{iclr2023_conference}
\bibliographystyle{iclr2023_conference}

\newpage
\appendix
\section{Experimental Setup}
\label{sec:experimental_setup}
We provide some additional information about the experimental setup for reproducing our results.

\begin{table}[h!]
    \caption{The architecture of Low-Resolution Decoder in pytorch style. $H, W$ are height and weight of target image. Here the size of the latent codes are assumed to be $12 \times H/8 \times W/8$.}
	\small
	\centering
	\begin{tabular}{ll}	
	    \toprule
		{\textbf{Output Size}} & {\textbf{Layers}} \\
		\hline
		{$12 \times H/8 \times W/8$} & {Latent Codes} \\
		{$9\times H/4 \times W/4$} & {\texttt{ConvTranspose2d}(in\_channels=12, out\_channels=9, kernel\_size=2, stride=2)} \\
		{$6\times H/2 \times W/2$} & {\texttt{ConvTranspose2d}(in\_channels=9, out\_channels=6, kernel\_size=2, stride=2)} \\
		{$3\times H \times W$} & {\texttt{ConvTranspose2d}(in\_channels=6, out\_channels=3, kernel\_size=2, stride=2)} \\
		{$3\times H \times W$} & {\texttt{Sigmoid}} \\
		\bottomrule
	\end{tabular}
	\label{tbl:low_resolution_decoder}
\end{table}

\begin{table}[h!]
    \caption{The architecture of High-Resolution Decoder in pytorch style. $H, W$ are height and weight of target image. Here the size of the latent codes are assumed to be $12 \times H/4 \times W/4$.}
	\small
	\centering
	\begin{tabular}{ll}
	    \toprule
		{\textbf{Output Size}} & {\textbf{Layers}} \\
		\hline
		{$12 \times H/4 \times W/4$} & {Latent Codes} \\
		{$6\times H/2 \times W/2$} & {\texttt{ConvTranspose2d}(in\_channels=12, out\_channels=6, kernel\_size=2, stride=2)} \\
		{$3\times H \times W$} & {\texttt{ConvTranspose2d}(in\_channels=6, out\_channels=3, kernel\_size=2, stride=2)} \\
		{$3\times H \times W$} & {\texttt{Sigmoid}} \\
		\bottomrule
	\end{tabular}
	\label{tbl:high_resolution_decoder}
\end{table}

\begin{table}[h!]
\caption{The hyperparameter configurations of KFS for each setting. ``I/C'' denotes ``Image / Class''. ``Low-R'' refers to the decoder in Table \ref{tbl:low_resolution_decoder}, and ``High-R'' refers to the decoder in Table. \ref{tbl:high_resolution_decoder}  }
	\small
	\centering
	\begin{tabular}{c|c|ccccc}
	    \toprule
	    {Dataset} & {I/C} & {Code Shape} & {\# of Code} & {Decoder Type} & {\# of Decoder} & {Over-Param. (\%)} \\
	    \hline
		\multirow{3}{*}{SVHN}
		& $1$
		& $12\times4\times4$
		& $13$
		& Low-R.
		& $8$
		& $0.47\%$
		\\
		& $10$
		& $12\times4\times4$
		& $160$
		& Low-R.
		& $12$
		& $2.88\%$
		\\
		& 50
		& $12\times8\times8$
		& $200$
		& High-R.
		& $16$
		& $0.88\%$
		\\
		\hline
		\multirow{3}{*}{CIFAR10}
		& $1$
		& $12\times4\times4$
		& $13$
		& Low-R.
		& $8$
		& $0.47\%$
		\\
		& $10$
		& $12\times4\times4$
		& $160$
		& Low-R.
		& $12$
		& $2.88\%$
		\\
		& 50
		& $12\times8\times8$
		& $200$
		& High-R.
		& $16$
		& $0.88\%$
		\\
		\hline
		\multirow{2}{*}{CIFAR100}
		& $1$
		& $12\times4\times4$
		& $16$
		& Low-R.
		& $8$
		& $1.92\%$
		\\
		& $10$
		& $12\times4\times4$
		& $160$
		& Low-R.
		& $12$
		& $0.29\%$
		\\
		\hline
		\multirow{2}{*}{TinyImageNet}
		& $1$
		& $12\times8\times8$
		& $16$
		& Low-R.
		& $8$
		& $0.24\%$
		\\
		& $10$
		& $12\times16\times16$
		& $64$
		& High-R.
		& $16$
		& $0.04\%$
		\\
		\bottomrule
	\end{tabular}
	\label{tbl:main_hyperparameter}
\end{table}

\begin{itemize}
    \item See Table \ref{tbl:low_resolution_decoder}, \ref{tbl:high_resolution_decoder} for detailed implementations of decoders.
    
    \item See Table \ref{tbl:main_hyperparameter} for code shape, \# of code, decoder type, \# of decoder, and corresponding proportion of over-parameterization on each setting.
    
    \item We pre-train a single decoder using autoencoding objective for 2,000 steps with 256 mini-batch. We use Adam optimizer \citep{adam} with constant learning rate of $0.01$. After pre-training, the parameter of pre-trained decoder is copied to all the others.
    
    \item We train decoders and codes using the full distribution matching objective in Eq.~\eqref{eq:dm} for 20,000 steps. We also use Adam optimizer with constant learning rate of 0.01 and 0.1 for decoders and latent codes, respectively.
    
    \item We train classification models for 200 epochs with 256 mini-batch on our condensed datasets. We use SGD with momentum optimizer, where the initial learning rate, momentum, and weight decay are set to 0.01, 0.9, and 0.0005, respectively. We decay the learning rate by factor of 0.2 for two times at 133 and 166-th epoch.
    
\end{itemize}

\newpage
\section{Derivation of Eq.~\eqref{eq:bias}}
\label{sec:bias}
For notational brevity, let $L_c(\Theta,\phi) := \frac{1}{2} \left\| \frac{1}{N} \sum_{n=1}^{N} g(x_{c,n}) - \frac{1}{DM}\sum_{d=1}^D \sum_{m=1}^{M} g(f_{c,m,d}) \right\|_2^2$ where $f_{c,m,d} := f(\theta_{c,m};\phi_d)$. We first compute $\nabla_{\Theta,\phi} L_c(\Theta,\phi)$.
\begin{align}
    &\nabla_{\Theta,\phi} L_c(\Theta,\phi) \nonumber\\
    & = \nabla_{\Theta,\phi} \frac{1}{2}\left\| \frac{1}{N} \sum_{n=1}^{N} g(x_{c,n}) - \frac{1}{DM}\sum_{d=1}^D \sum_{m=1}^{M} g(f_{c,m,d}) \right\|_2^2 \nonumber \\
    &=  -\frac{1}{NDM} \sum_{n=1}^N \sum_{d=1}^D \sum_{m=1}^M \nabla_{\Theta,\phi} g(f_{c,m,d}) \tran g(x_{c,n})  + \frac{1}{D^2 M^2} \sum_{d=1}^D \sum_{d'=1}^D \sum_{m=1}^M \sum_{m'=1}^M \nabla_{\Theta,\phi} g(f_{c,m',d'}) \tran g(f_{c,m,d})
    \label{eq:bias_gt}
\end{align}
On the other hand, suppose we sample $m \sim \text{Uniform}(1,M)$ and $d \sim \text{Uniform}(1,D)$, then
\begin{align}
    &\mathbb{E}_{m,d}[\nabla_{\Theta,\phi}\hat{L}_c(\Theta,\phi)] \nonumber \\
    & = \mathbb{E}_{m,d}\left[ \nabla_{\Theta,\phi} \frac{1}{2}\left\| \frac{1}{N} \sum_{n=1}^{N} g(x_{c,n}) - g(f_{c,m,d}) \right\|_2^2\right] \nonumber \\
    & = \mathbb{E}_{m,d}\left[ -\frac{1}{N} \sum_{n=1}^N g(x_{c,n})\tran \nabla_{\Theta,\phi} g(f_{c,m,d}) + \nabla_{\Theta,\phi} g(f_{c,m,d}) \tran g(f_{c,m,d})  \right] \nonumber \\
    &= -\frac{1}{NDM} \sum_{n=1}^N \sum_{d=1}^D \sum_{m=1}^M \nabla_{\Theta,\phi} g(f_{c,m,d})\tran g(x_{c,n}) + \frac{1}{D M} \sum_{d=1}^D \sum_{m=1}^M \nabla_{\Theta,\phi} g(f_{c,m,d}) \tran g(f_{c,m,d}) ) 
    \label{eq:bias_samp}
\end{align}
Therefore, subtracting Eq.~\eqref{eq:bias_gt} from Eq.~\eqref{eq:bias_samp}, we have
\begin{align}
    &\mathbb{E}_{m,d}[\nabla_{\Theta,\phi}\hat{L}(\Theta,\phi)] - \nabla_{\Theta,\phi}L(\Theta,\phi) \nonumber \\
    &= \frac{1}{C}\sum_{c=1}^C \left(\mathbb{E}_{m,d}[\nabla_{\Theta,\phi}\hat{L}_c(\Theta,\phi)] - \nabla_{\Theta,\phi}L_c(\Theta,\phi)\right) \nonumber \\
    &= \nabla_{\Theta, \phi} \frac{1}{C}\sum_{c=1}^C \frac{1}{2} \Bigg\{
    \frac{1}{DM} \sum_{d=1}^D \sum_{m=1}^M g(f_{c,m,d})\tran g(f_{c,m,d}) 
    - \frac{1}{D^2M^2} \sum_{d=1}^D\sum_{d'=1}^D\sum_{m=1}^M\sum_{m'=1}^M g(f_{c,m,d})\tran g(f_{c,m',d'})
    \Bigg\}. \nonumber
\end{align}

\newpage
\section{Derivation of Eq.~\eqref{eq:variance}}
\label{sec:variance}

We want to compute
\begin{align}
    &\text{Var}_n(\nabla_{\Theta,\phi}\tilde{L}(\Theta,\phi)) \nonumber \\
    &= \mathbb{E}_n\left[\nabla_{\Theta,\phi}\tilde{L}(\Theta,\phi) \nabla_{\Theta,\phi}\tilde{L}(\Theta,\phi)\tran\right] - \mathbb{E}_n\left[ \nabla_{\Theta,\phi}\tilde{L}(\Theta,\phi) \right] \mathbb{E}_n\left[ \nabla_{\Theta,\phi}\tilde{L}(\Theta,\phi) \right]\tran \nonumber
\end{align}
Suppose we sample $n \sim \text{Uniform}(1,N)$ for each class $c$ independently. Then we have
\begin{align}
    &\mathbb{E}_n\left[\nabla_{\Theta,\phi}\tilde{L}(\Theta,\phi) \right] \nonumber \\
    &= \mathbb{E}_n \left[ \nabla_{\Theta,\phi} \frac{1}{C}\sum_{c=1}^C \frac{1}{2}\left\| g(x_{c,n}) - \frac{1}{DM}\sum_{d=1}^D \sum_{m=1}^{M} g(f_{c,m,d}) \right\|_2^2 \right] \nonumber \\
    &= \mathbb{E}_n\left[\frac{1}{C}\sum_{c=1}^C \left( -\frac{1}{DM}\sum_{d=1}^D \sum_{m=1}^M  \nabla_{\Theta,\phi} g(f_{c,m,d})\tran g(x_{c,n}) + \frac{1}{D^2 M^2} \sum_{d=1}^D \sum_{d'=1}^D \sum_{m=1}^M \sum_{m'=1}^M  \nabla_{\Theta,\phi} g(f_{c,m',d'}) \tran g(f_{c,m,d})  \right)\right] \nonumber \\
    &= \frac{1}{C}\sum_{c=1}^C  \frac{1}{DM} \sum_{d=1}^D \sum_{m=1}^M \nabla_{\Theta,\phi} g(f_{c,m,d}) \tran \left( - \mathbb{E}_n[g(x_{c,n})] + \frac{1}{DM} \sum_{d'=1}^D \sum_{m'=1}^M g(f_{c,m',d'}) \right) \nonumber \\
    &= \frac{1}{C}\sum_{c=1}^C  V_c \tran \left( - \mathbb{E}_n [g(x_{c,n})] + \frac{1}{DM} \sum_{d'=1}^D \sum_{m'=1}^M g(f_{c,m',d'}) \right) \nonumber
\end{align}
where $V_c := \frac{1}{DM} \sum_{d=1}^D \sum_{m=1}^M \nabla_{\Theta,\phi} g(f_{c,m,d})$. Therefore,
\begin{align}
    &\mathbb{E}_n\left[\nabla_{\Theta,\phi}\tilde{L}(\Theta,\phi) \right] \mathbb{E}_n\left[\nabla_{\Theta,\phi}\tilde{L}(\Theta,\phi) \right]\tran \nonumber \\
    &= \frac{1}{C}\sum_{c=1}^C  V_c \tran \left( -  \mathbb{E}_n [g(x_{c,n})] + \frac{1}{DM} \sum_{d=1}^D \sum_{m=1}^M g(f_{c,m,d}) \right) \cdot \nonumber \\
    &\qquad \frac{1}{C}\sum_{c'=1}^C \left( -  \mathbb{E}_{n} [g(x_{c',n})] + \frac{1}{DM} \sum_{d'=1}^D \sum_{m'=1}^M g(f_{c',m',d'}) \right) \tran V_{c'} \nonumber \\
    &= \frac{1}{C^2} \sum_{c=1}^C \sum_{c'=1}^C V_c\tran \bigg(  \mathbb{E}_n [g(x_{c,n})] \mathbb{E}_n [g(x_{c',n})]\tran - \frac{2}{DM}  \sum_{d=1}^D \sum_{m=1}^M  \mathbb{E}_n [g(x_{c,n})]\tran g(f_{c',m,d}) \nonumber \\
    &\qquad\qquad\qquad\qquad\ \ + \frac{1}{D^2 M^2} \sum_{d=1}^D \sum_{d'=1}^D \sum_{m=1}^M \sum_{m'=1}^M g(f_{c,m,d})\tran g(f_{c',m',d'}) \bigg) V_{c'}
    \label{eq:var_second}
\end{align}
On the other hand, we have
\begin{align}
    &\mathbb{E}_n\left[\nabla_{\Theta,\phi}\tilde{L}(\Theta,\phi) \nabla_{\Theta,\phi}\tilde{L}(\Theta,\phi)\tran\right] \nonumber \\
    &= \mathbb{E}_n \Bigg[ \Bigg\{\frac{1}{C} \sum_{c=1}^C \frac{1}{DM} \sum_{d=1}^D \sum_{m=1}^M \nabla_{\Theta,\phi} g(f_{c,m,d}) \tran \left( -g(x_{c,n}) + \frac{1}{DM} \sum_{d'=1}^D \sum_{m'=1}^M g(f_{c,m',d'}) \right)\Bigg\} \cdot \nonumber \\
    &\qquad\quad\  \Bigg\{\frac{1}{C} \sum_{c=1}^C \frac{1}{DM} \sum_{d=1}^D \sum_{m=1}^M \nabla_{\Theta,\phi} g(f_{c,m,d}) \tran \left( -g(x_{c,n}) + \frac{1}{DM} \sum_{d'=1}^D \sum_{m'=1}^M g(f_{c,m',d'}) \right)\Bigg\}\tran \Bigg]\nonumber \\
    &= \frac{1}{C^2} \sum_{c=1}^{C} \sum_{c'=1}^C V_c\tran \Bigg(  \mathbb{E}_n [g(x_{c,n}) g(x_{c',n})\tran] - \frac{2}{DM} \sum_{d=1}^D \sum_{m=1}^M  \mathbb{E}_n [g(x_{c,n})]\tran g(f_{c',m,d}) \nonumber \\
    &\qquad\qquad\qquad\qquad\ \ + \frac{1}{D^2 M^2} \sum_{d=1}^D \sum_{d'=1}^D \sum_{m=1}^M \sum_{m'=1}^M g(f_{c,m,d})\tran g(f_{c',m',d'}) \Bigg) V_{c'}
    \label{eq:var_first}
\end{align}
Subtracting Eq.~\eqref{eq:var_second} from Eq.~\eqref{eq:var_first}, we have
\begin{align}
    &\text{Var}_n(\nabla_{\Theta,\phi}\tilde{L}(\Theta,\phi)) \nonumber \\
    &= \frac{1}{C^2} \sum_{c=1}^C \sum_{c'=1}^C V_c\tran \Bigg( \mathbb{E}_n\left[ g(x_{c,n})g(x_{c',n})\right]\tran - \mathbb{E}_n[g(x_{c,n})]\mathbb{E}_n[g(x_{c',n})]\tran \Bigg) V_{c'} \nonumber \\
    &= \frac{1}{C^2} \sum_{c = 1}^C V_c\tran \left( \mathbb{E}_n[g(x_{c,n})g(x_{c,n})\tran] - \mathbb{E}_n[g(x_{c,n})]\mathbb{E}_n[g(x_{c,n})]\tran \right) V_{c} \nonumber \\
    &= \frac{1}{C^2}\sum_{c=1}^C V_c \left\{
    \frac{1}{N} \sum_{n=1}^N g(x_{c,n}) g(x_{c,n} )\tran
    - \left(\frac{1}{N} \sum_{n=1}^N  g(x_{c,n})\right) \left(\frac{1}{N} \sum_{n=1}^N  g(x_{c,n})\right)\tran
    \right\} V_c\tran
    \nonumber
\end{align}
because $g(x_{c,n})$ and $g(x_{c',n})$ are independent of each other as we sample $n$ independently for each class.

\end{document}